\documentclass[a4paper,fleqn]{cas-dc}

\usepackage[numbers]{natbib}

\usepackage{multirow}
\usepackage{diagbox}
\usepackage{graphicx}
\usepackage{subfigure}
\usepackage[linesnumbered, ruled]{algorithm2e}
\usepackage{pifont}  % 加载 pifont 包
\usepackage{xcolor}  % 加载 xcolor 包

\begin{document}
\let\WriteBookmarks\relax
\def\floatpagepagefraction{1}
\def\textpagefraction{.001}
\shorttitle{NQKV}

\shortauthors{Zhihang Cai et~al.}

\title[mode = title]{NQKV: A KV Cache Quantization Scheme Based on Normal Distribution Characteristics}  

\address[1]{School of Computer Science and Technology, Xi'an Jiaotong University, Xi'an 710049}

\author[1]{Zhihang Cai}
\credit{Conceptualization, Methodology, Software, Writing - original draft}

\author[1]{Xingjun Zhang}[orcid=0000-0003-1434-7016]
\cormark[1]
\ead{xjzhang@xjtu.edu.cn}
\credit{Resources, Funding acquisition, Project administration, Supervision}

\author[1]{Zhendong Tan}
\credit{Writing - review \& editing, Validation}

\author[1]{Zheng Wei}[orcid=0000-0002-2293-5427]
\credit{Writing - review \& editing, Validation}

\cortext[1]{Corresponding author}

\begin{abstract}Large Language Models (LLMs) have demonstrated remarkable proficiency across a wide range of tasks. However, LLMs often require larger batch sizes to enhance throughput or longer context lengths to meet task demands, which significantly increases the memory resource consumption of the Key-Value (KV) cache during inference, becoming a major bottleneck in LLM deployment. To address this issue, quantization is a common and straightforward approach. Currently, quantization methods for activations are limited to 8-bit, and quantization to even lower bits can lead to substantial accuracy drops. To further save space by quantizing the KV cache to even lower bits, we analyzed the element distribution of the KV cache and designed the NQKV algorithm. Since the elements within each block of the KV cache follow a normal distribution, NQKV employs per-block quantile quantization to achieve information-theoretically optimal quantization error. Without significantly compromising model output quality, NQKV enables the OPT model to perform inference with an 2× larger batch size or a 4× longer context length, and it improves throughput by 9.3× compared to when the KV cache is not used.
\end{abstract}


\maketitle

\section{Introduction}
\label{}

\begin{figure}[b]
    \centering
    \includegraphics[width=0.9\linewidth]{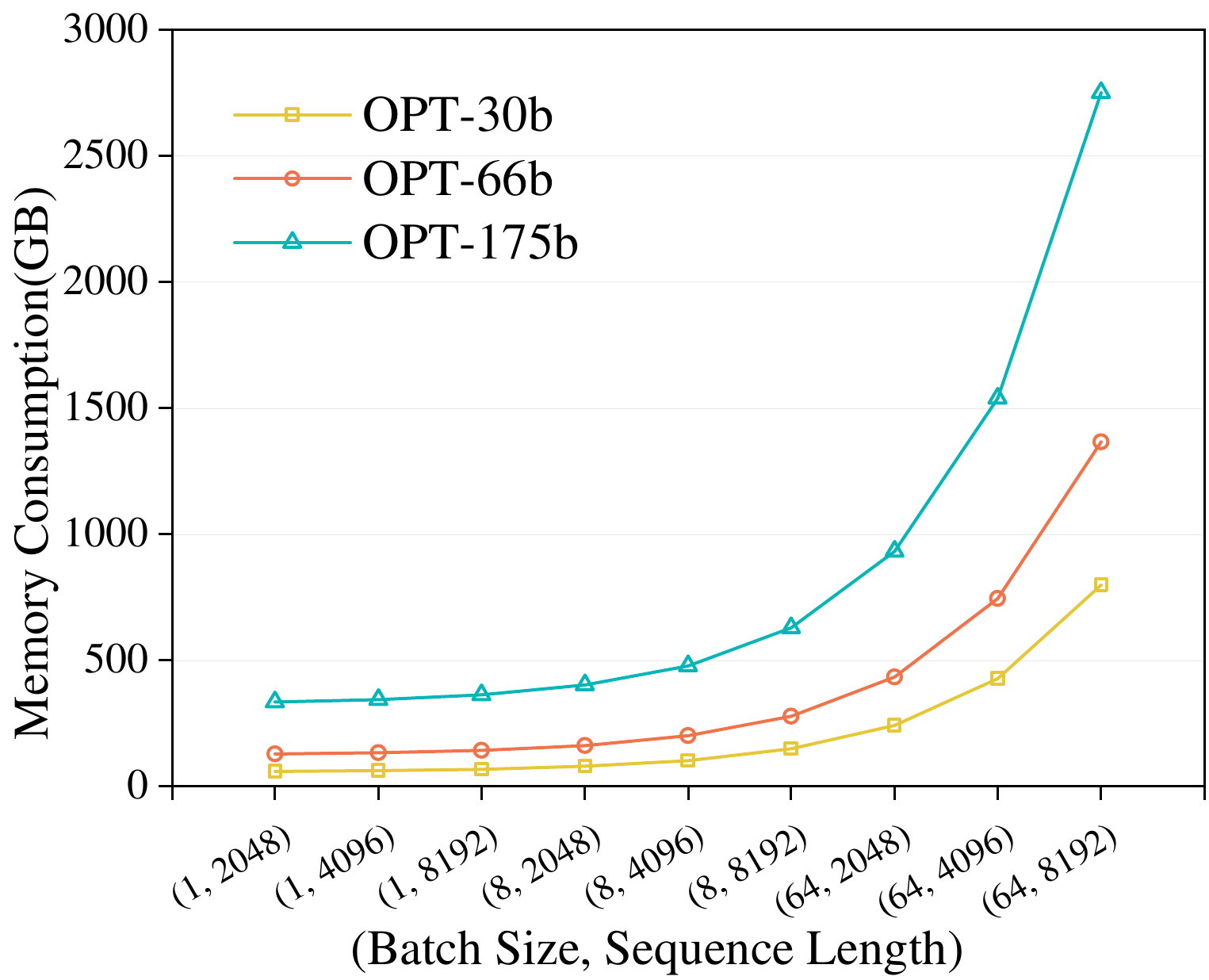}
    \caption{The memory comsumption of OPT models in different scales under various batch size and sequence length configurations.}
    \label{fig.model_size}
\end{figure}

\begin{figure}[t]
    \centering
    \includegraphics[width=1.0\linewidth]{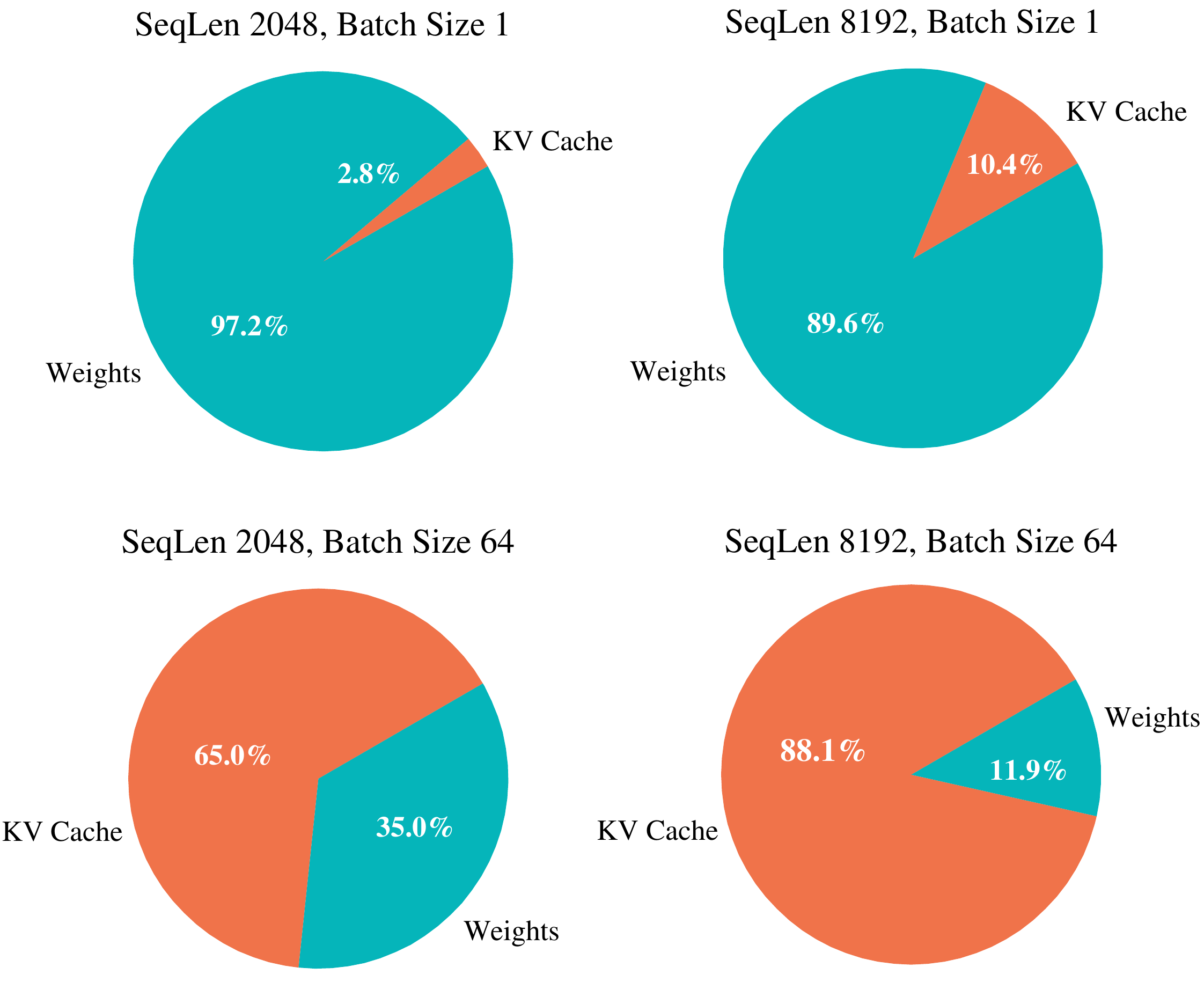}
    \caption{The memory usage percentages of different components during inference for the OPT-175B model. As the batch size and sequence length increase, the memory space allocated to the KV cache ignificantly increases.}
    \label{fig.KV_proportion}
\end{figure}

Large Language Models (LLMs) have shown impressive performance across a wide range of tasks \cite{brown2020language, yuan2023llm, zhang2022opt}.  As LLMs are tasked with increasingly complex problems, they often require larger batch sizes to maximize GPU utilization and throughput, or longer context lengths to generate higher quality and more relevant output. However, large batch sizes and long context lengths significantly increase the memory footprint of LLMs during inference, posing new challenges for deploying and running LLMs \cite{chen2023longlora}. As shown in Fig.~\ref{fig.model_size}, the GPU memory usage during LLM inference time increases sharply with larger batch sizes and longer sequence lengths. This effect is particularly pronounced for models with a greater number of parameters. In this scenario, compared to the model weights, the KV cache, which stores the keys and values of the attention mechanism during inference to prevent redundant calculations, occupies the majority of the GPU memory space.  We present the proportion of GPU memory usage by the KV cache under different batch sizes and sequence lengths in Fig.~\ref{fig.KV_proportion}. For example,  the proportion of  KV cache memory usage during inference for the OPT-175B model reaches 83.78\% when the batch size is 64 and the sequence length is 8192. Specifically, the KV cache would occupy 2.3TB of space, which is seven times the size of the model's own parameters. In such cases, the KV cache becomes the primary bottleneck for deploying and performing inference on large language models \cite{yuan2023rptq}. Therefore, reducing the memory overhead of the KV cache while maintaining model accuracy is an important way to lower the deployment costs of large language models.

% \begin{table}
% \centering
% \caption{The memory usage percentages of different components during inference for the OPT-175B model. The 'Weight' column represents the percentage of memory used by the model's weights, 'K/V' denotes the percentage taken up by the Key-Value (KV) cache, and 'Dynamic' refers to the memory used by other dynamic components. As the batch size and sequence length increase, the memory space allocated to the KV cache significantly increases.}
% \label{table:footprInt}
%     \scalebox{0.9}{
%         \begin{tabular}{lcccc}
%             \toprule
%             Batch Size & Sequence Length & Weight & K/V & Dynamic \\ 
%             \midrule
            
%             1 & 2048 & 97.18\% & 2.68\% & 0.14\% \\ 
%             ~ & 8192 & 89.59\% & 9.89\% & 0.52\% \\ 
%             64 & 2048 & 34.98\% & 61.80\% & 3.22\% \\ 
%             ~ & 8192 & 11.85\% & 83.78\% & 4.37\% \\ 
%             \bottomrule
%         \end{tabular}
%     }
% \end{table}

Currently, there are several approaches to reducing the memory footprint of the KV cache in resource-constrained scenarios to improve memory efficiency. Some efforts attempt to address the issue at the system level. Offloading \cite{sheng2023flexgen} is a practical method to alleviate memory pressure during model inference when dealing with excessively long contexts. Although offloading can effectively reduce memory usage, it poses a complex challenge due to its high dependency on data transmission bandwidth. There are also efforts that attempt to incorporate virtual memory and paging techniques into the attention mechanism \cite{kwon2023efficient}. Additionally, some methods focus on reducing the number of heads in the KV cache, such as multi-query attention \cite{shazeer2019fast} and multi-group attention \cite{ainslie2023gqa}. However, these methods modify the model's architecture, requiring subsequent retraining or fine-tuning of the model. Other methods employ cache eviction strategies to evict less important tokens from the KV cache \cite{zhang2024h2o}. Each of these methods has its own challenges, including complex implementation and difficulty in integrating with existing models.

Quantization offers a promising approach to reducing the cost of LLMs. By quantizing the KV cache into a lower-bit data type, we can reduce memory requirements. For example, a 8-bit quantization of KV cache can reduce memory usage by half, while a 4-bit quantization would result in a memory space occupancy that is only one quarter of the original. There are numerous methods for quantizing weights \cite{frantar2022gptq}, \cite{dettmers2023spqr} and both weights and activations \cite{xiao2023smoothquant}, \cite{yao2022zeroquant}, \cite{dettmers2022gpt3}. However, these methods can not be directly used to quantize the KV cache for three reasons. Firstly, current quantization methods for activations struggle to maintain relatively low model accuracy loss at 4 bits \cite{dettmers2023case}. Secondly, these methods also quantize the weights, but in scenarios where the batch size is very large and the sequence length is very long, the benefits of quantizing the weights are minimal and can lead to an accuracy drop. An even more challenging issue is that, due to the streaming nature of KV cache, existing  methods cannot be directly applied to the KV cache. In this paper, we propose NQKV, which quantizes KV cache based on its normal distribution characteristics and uses data types that better align with the normal distribution. NQKV is based on the following insights:

\begin{itemize}
    \item In the transformer \cite{vaswani2017attention} architecture, the elements of keys and values in the decoder layer follow a normal distribution. Most quantization methods currently use integer (int) as the data type \cite{xiao2023smoothquant, frantar2022gptq, dettmers2022gpt3}. To leverage the normal distribution characteristics of the data, there are also quantization methods that use floating-point (float) data types \cite{kuzmin2022Fp8, zhang2023Integer}. By using data types that are closer to the normal distribution, it is possible to further reduce quantization error.

    \item The token dimension, when partitioned by block size, results in blocks that still conform to the normal distribution. We have observed that keys and values adhere to the normal distribution along the token dimension, allowing for per-token quantization. Furthermore, if each token is divided into blocks of a certain block size, the resulting tensors also conform to the normal distribution. This suggests that we can perform quantization at an even finer granularity beyond per-token quantization by using data types that align with the normal distribution to quantize each block. This approach confines quantization error within a block, preventing it from propagating across the entire token.
    
    \item The KV cache has a streaming nature. In the generative inference of LLMs, the KV cache stores all the keys and values computed by the attention mechanism in previous calculations. When generating new tokens, these cached values can be reused, thus avoiding redundant computations. After the new key and value tensors for the newly generated tokens are computed, they are appended directly to the end of the KV cache. During this process, the old keys and values remain stored in the KV cache and do not change over time; that is, the KV cache is append-only. This characteristic of the KV cache means that in addition to computational data types like int and float, we can also use other storage data 
    types such as NormalFloat \cite{dettmers2024qlora}.
  
    \item It is more appropriate to quantize the KV cache along the token dimension rather than the channel dimension. In text generation tasks, this ensures that  newly generated tokens will not affect the quantization of other tokens. Furthermore, after quantization, newly generated keys and values can be directly appended to the end of the KV cache. This aligns with the streaming nature of KV cache.

\end{itemize}
 % NQKV quantizes keys and values in token dimension. 
 % During the prefill phase, the generated keys and values are first quantized to 4-bit precision and then stored in the KV cache. In the subsequent decoding phase, the KV cache is first dequantized and then concatenated with the newly generated keys and values for the subsequent computation steps of the attention mechanism. After that, the newly generated keys and values are quantized to 4-bit precision, and the resulting values are directly appended to the end of the KV cache.

 % If the token dimensions of the keys and values participating in matrix multiplication are not multiples of 16, NQKV will perform padding, a
 
Inspired by these insights, we propose NQKV, a KV cache quantization method based on normal distribution. NQKV uses storage data types such as Normal Float \cite{dettmers2024qlora} rather than computational data types to represent quantized values. With a limited number of bits, storage data types allow for more flexible data point values. Therefore, we can select quantile points from a normal distribution as data points to minimize quantization error.  NQKV divides each token into several blocks based on a specified block size and quantizes each block separately. This not only utilizes the normal distribution properties of keys and values, but also limits quantization errors within a single block without spreading across the entire token. To accelerate KV cache quantization, NQKV employs padding techniques, allowing for the use of more efficient BMM kernels for matrix multiplication operations. Our contributions are summarized as follows:

\begin{itemize}
    \item \textbf{Extensive analysis of the
    element distribution within KV cache.} We found that both
    within individual tokens and within individual blocks, the
    elements follow a normal distribution. Our observations suggest that using data types whose data points follow a normal distribution for per-block quantization of KV cache.

    \item \textbf{A new 4bit KV cache quantization algorithm without any finetuning.} Based on the normal distribution characteristics of the KV cache, we propose an algorithm specifically designed for quantizing the KV cache, called NQKV. This method is orthogonal to other advanced model quantization techniques or system level memory management strategies and can be used in combination with them.   
    
    \item \textbf{Quantizing the KV cache to 4 bits with minimal accuracy drop.} Our experiments demonstrate that NQKV has a negligible impact on the model's accuracy. It enables an 2× larger batch size or 4× longer sequence length for inference when the KV cache is enabled, and it improves throughput by 9.3× compared to not using the KV cache.
\end{itemize}

\section{Related Work}

\paragraph{\textbf{Weight-only quantization.}} Quantization is a commonly used technique to compress model size and reduce model inference overhead \cite{nagel2021white, zhu2023survey, han2015deep}. Some works focus on quantizing model weights, representing weights with lower bit data types to decrease model size. In scenarios with small batch size and sequence length, model weights are the primary source of memory consumption, thus these methods can effectively reduce the model size. GPTQ \cite{frantar2022gptq} utilizes approximate second-order information to quantize model weights with negligible impact on model performance. AWQ \cite{lin2023awq} perceives the importance of weights based on the magnitude of activation values rather than the weights themselves, and further protects important weights, successfully quantizing weights to 4 bits and 3 bits. SpQR \cite{dettmers2023spqr} observes that outliers in weights are the main cause of quantization difficulty. Therefore, SpQR can identify outliers in weights and store them using higher precision data types to reduce the accuracy drop caused by quantization. SqueezeLLM \cite{kim2023squeezellm} leverages a second-order information driven strategy to search for the optimal bit precision, while also encoding outliers in a sparse format to mitigate quantization errors. These methods are orthogonal to our approach, as our method only operates on activations without involving weights, and thus does not conflict with these methods in implementation.

\paragraph{\textbf{Weight-activation quantization.}} If only the weights are quantized, the model still uses 16-bit floating point operations during inference, and thus cannot effectively utilize efficient low-bit matrix multiplication kernels to enhance computational speed and reduce inference latency. To address this problem and further reduce LLM's memory footpring, some works simultaneously quantize weights and activation values. SmoothQuant \cite{xiao2023smoothquant} quantizes both weights and activations to INT8. It has been observed that weights are relatively easier to quantize compared to activations. SmoothQuant achieves smaller quantization errors on activations by transferring the quantization difficulty from activations to weights through a mathematically equivalent transformation. This approach allows SmoothQuant to quantize the KV cache to INT8. However, when attempting to push activations to 4-bit quantization, SmoothQuant experiences a significant drop in accuracy. Qdrop \cite{wei2022qdrop} pushes the limit of PTQ to the 2-bit activation for the first time. It accomplishes this by randomly dropping the quantization of activations during PTQ. Outlier Suppression+ \cite{wei2023outlier} finds that outliers are concentrated in specific channels and exhibit asymmetry across channels. It utilizes channel-wise shifting to eliminate this asymmetry characteristic. GPT3.int8() \cite{dettmers2022gpt3} reduces the difficulty of activations quantization through another approach: it uses FP16 to represent outliers in activation values and INT8 to represent other activation values. However, this implementation leads to increased inference latency, even exceeding that of FP16 models. Although these methods can be used to quantize the KV cache, they are not specifically designed for it and do not take its streaming nature into account. As a result, quantizing the KV cache to lower bit levels such as 4-bit can lead to a severe drop in accuracy. Therefore, there are still difficulties in pushing the quantization of KV cache to lower bit levels.

\paragraph{\textbf{KV cache-only quantization.}} Furthermore, there are some works specifically targeting KV cache quantization. Llm-qat \cite{liu2023llm} can quantize the KV cache to 4 bits, but it requires retraining or fine-tuning to maintain performance. This process is extremely costly for LLMs. Another concurrent work \cite{liukivi} observes the differences between key cache and value cache, and proposes per-token quantization for key cache and per-channel quantization for value cache. However, due to the streaming nature of KV cache, per-channel quantization cannot be directly applied to value cache, necessitating a specialized implementation. Additionally, this implementation cannot avoid a portion of value cache still needing to be represented in FP16 during inference.

\paragraph{\textbf{Memory-efficient system.}}In addition to quantization, other works attempt to address this problem from different perspectives. vLLM \cite{kwon2023vllm} and S3 \cite{jin2024s3} are system-level works. They integrate memory management strategies like PagedAttention or memory usage prediction to diminish the memory footprint of the KV cache. These methods not only alleviate memory requirements but also enhance model throughput. StreamingLLM \cite{xiao2023streaming} is built upon the insight of the "attention sink" phenomenon and retains only a small number of initial tokens to preserve performance. These methods are orthogonal to NQKV, and these improvements can also be leveraged to enhance the performance of our algorithm.

\begin{figure*}
    \centering
    \includegraphics[width=0.9\textwidth]{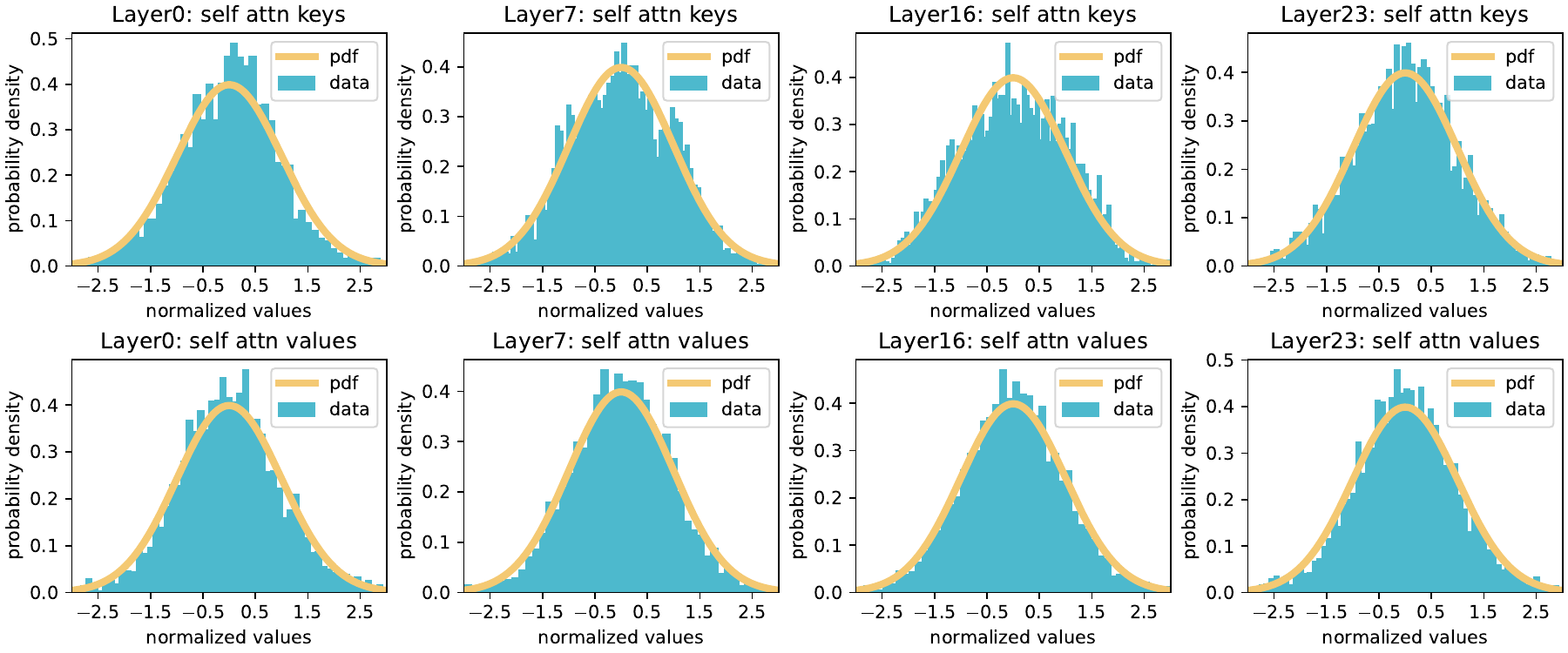}
    \caption{Demonstration of the data distribution of randomly selected tokens in OPT-6.7B decoder layers. Even if the data within each token follows a normal distribution, their standard deviations may differ. Therefore, we standardized the data to make their standard deviations equal to 1, allowing for easy comparison with the standard normal distribution. For ease of observation, we also plotted the probability density function curve of the standard normal distribution in the figure.}
    \label{fig.distribution}
\end{figure*}
%The Q-Q plot of a randomly selected token in OPT-6.7B model.
%The Q-Q plot of blocks in the token.
\begin{figure*}
    \subfigure[token]{
        \begin{minipage}[t]{0.225\linewidth}
            \centering
            \includegraphics[width=1.0\linewidth]{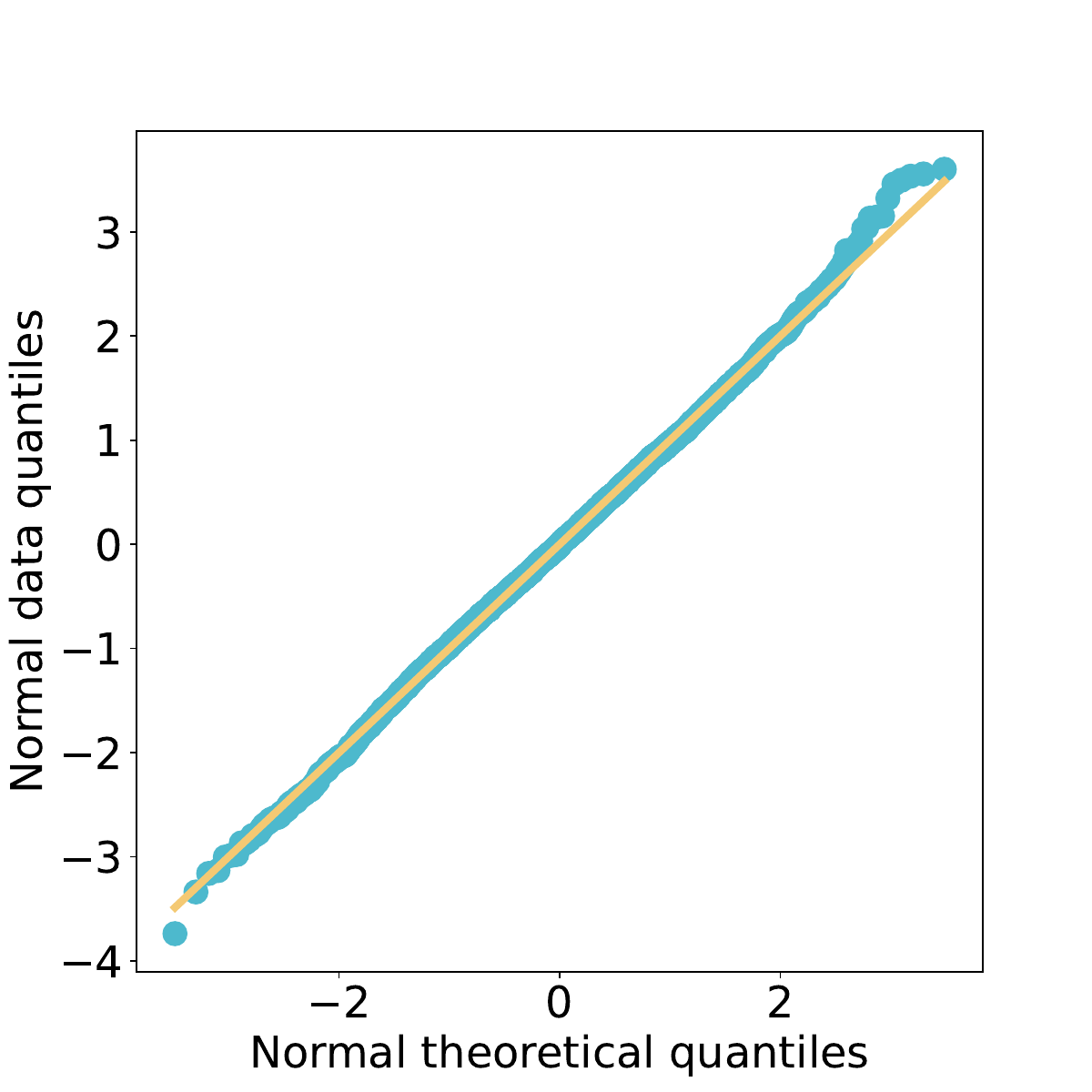} 
            \label{fig.qq.token}
        \end{minipage}
    }
    \subfigure[block1]{
        \begin{minipage}[t]{0.225\linewidth}
            \centering
            \includegraphics[width=1.0\linewidth]{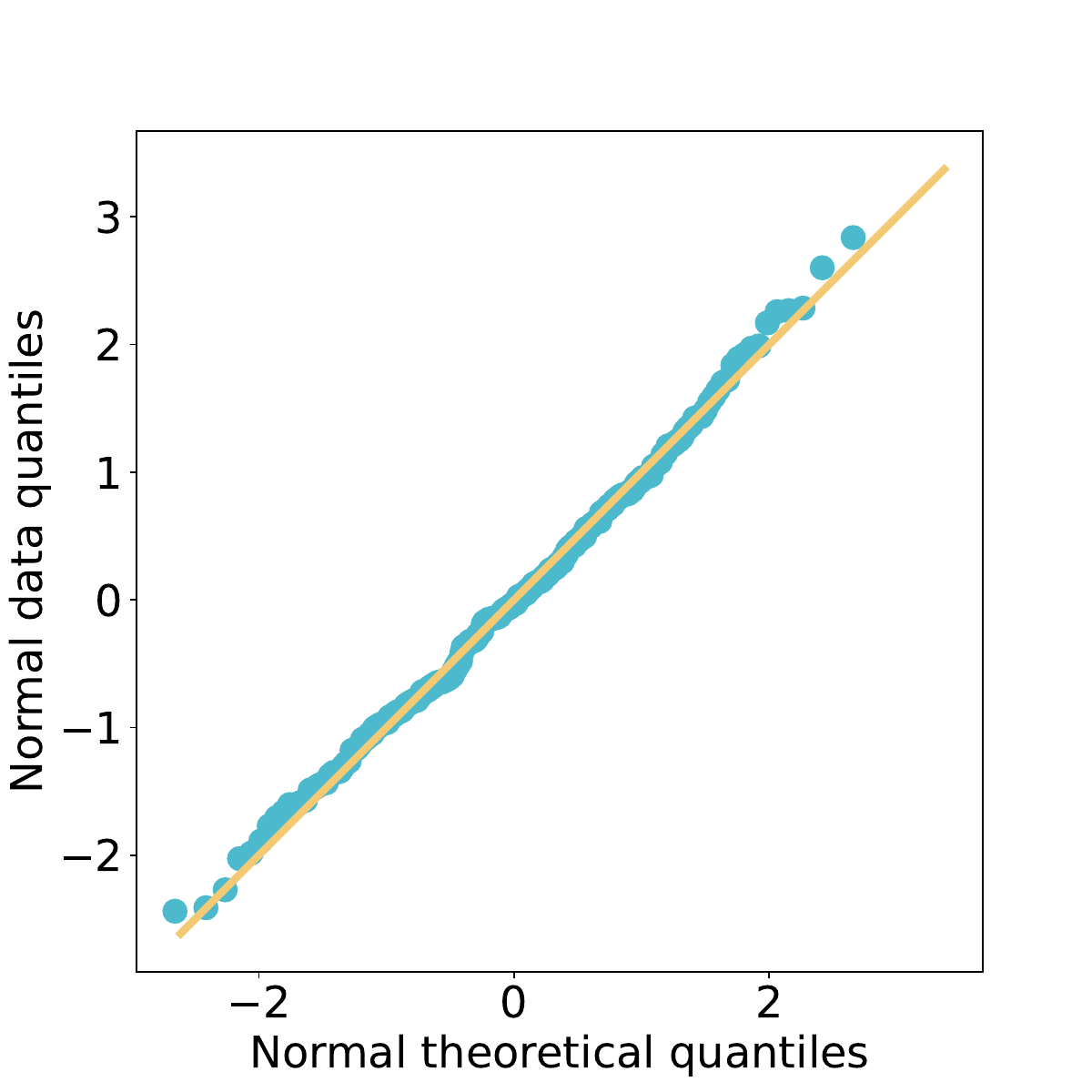}
                        \label{fig.qq.block1}
        \end{minipage}
    }
        \subfigure[block2]{
        \begin{minipage}[t]{0.225\linewidth}
            \centering
            \includegraphics[width=1.0\linewidth]{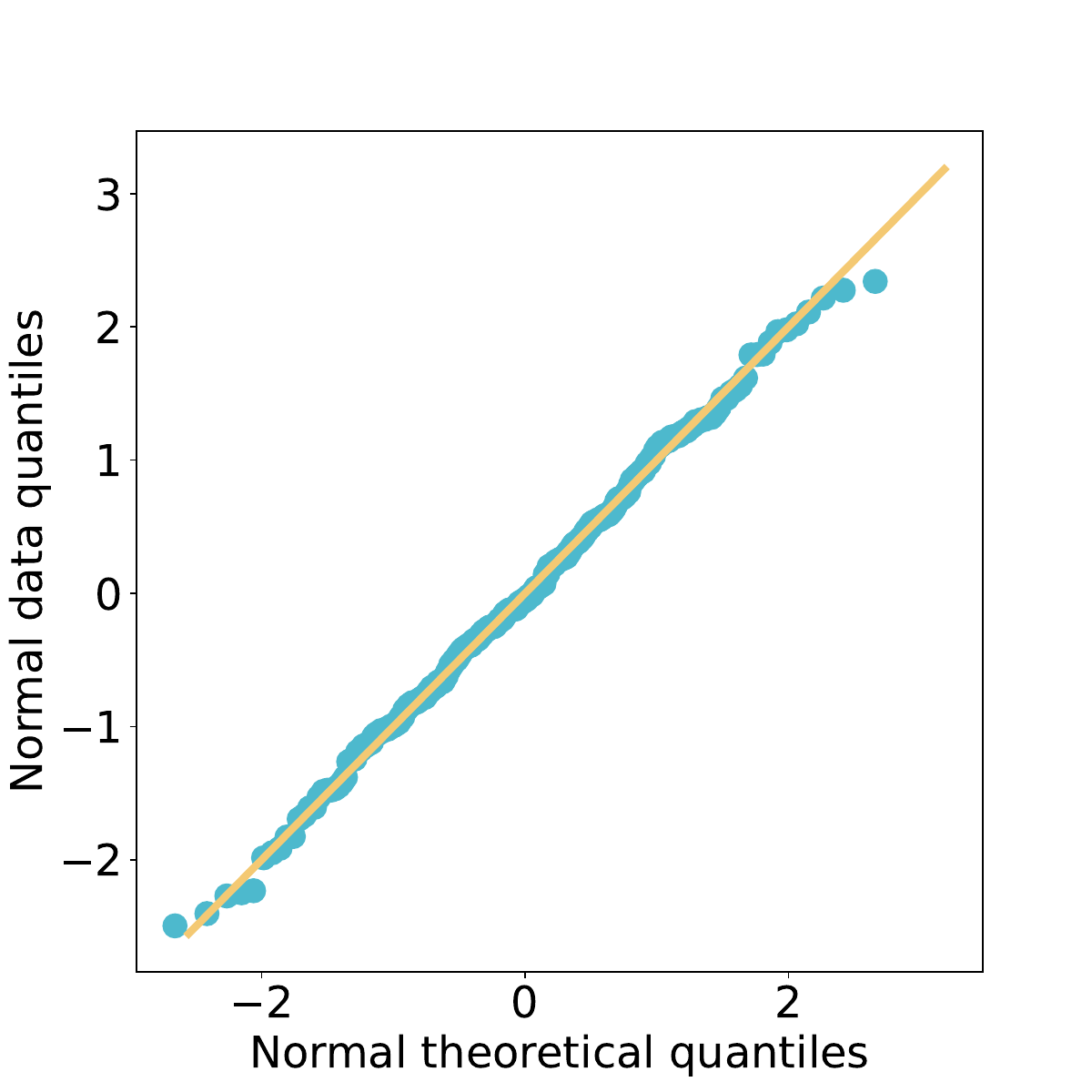}
                        \label{fig.qq.block2}
        \end{minipage}
    }
        \subfigure[block3]{
        \begin{minipage}[t]{0.225\linewidth}
            \centering
            \includegraphics[width=1.0\linewidth]{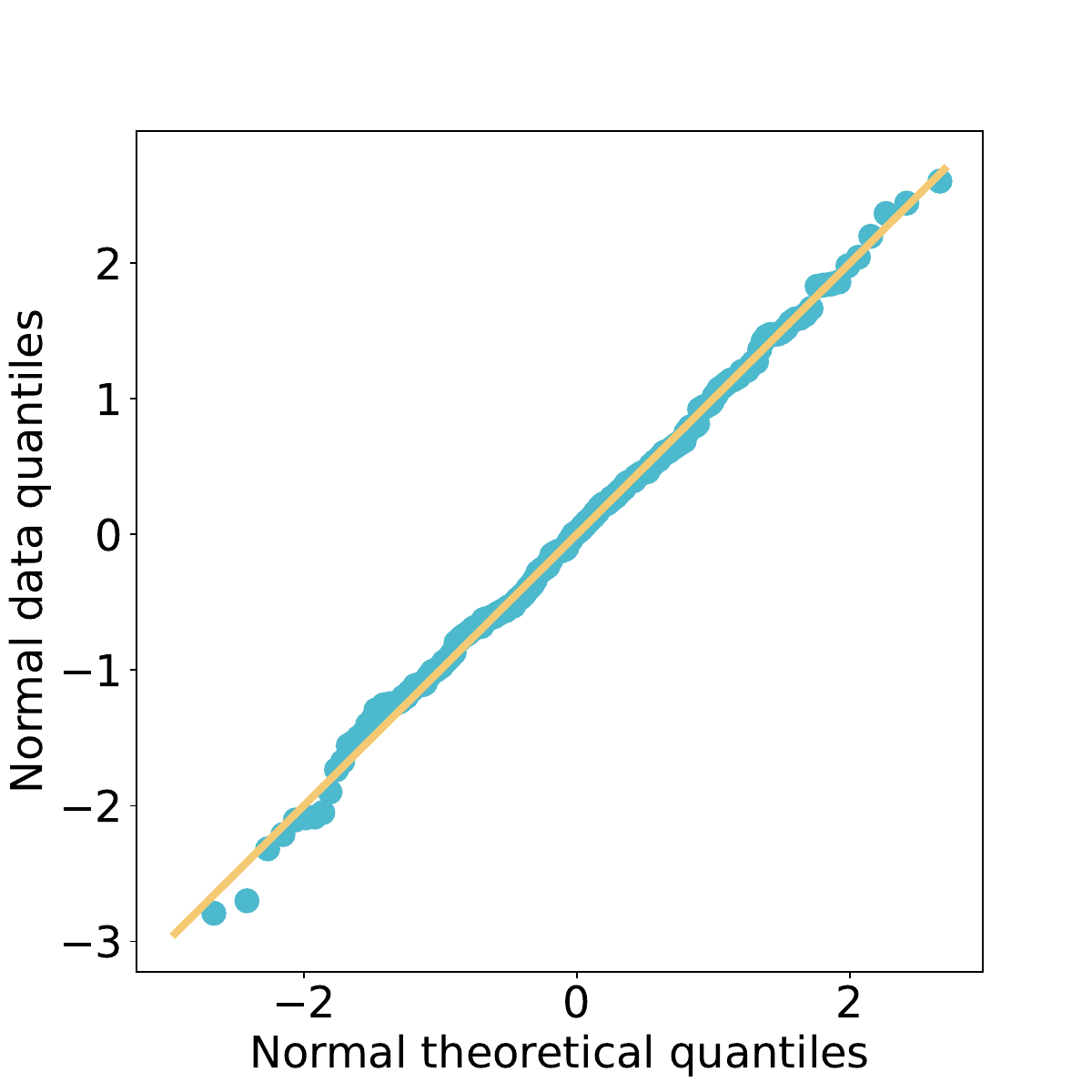}
                        \label{fig.qq.block3}
        \end{minipage}
    }
\centering
\caption{Quantile-Quantile plots of data distribution in tokens and blocks of the OPT-6.7B model. The hidden states size of OPT-6.7B is 4096. With a block size of 256, we can obtain 16 blocks. For the sake of demonstration, only the Quantile-Quantile plots of three of these blocks are shown here. The identity line $y=x$ represents the Q-Q plot of a standard normal distribution, while other data points are plotted based on the distribution of the data. If the data points approximately lie on the line $y=x$, it indicates that the two distributions being compared are similar, that is, the data follows a normal distribution.}
\label{fig.qq}
\end{figure*}

\section{Method}

In scenarios with large batch size and long context inference, we find that the memory storage occupied by KV cache significantly increases, becoming the main bottleneck for deploying LLM inference. To address this issue, quantization is a simple and effective method. It reduces the number of bits occupied by each activation, thereby reducing the overall memory space occupied by the KV cache. Although there are many methods for quantizing weights and activation values, they are not specifically tailored for KV cache and can only quantize the KV cache to a maximum of 8 bits. When quantized to 4 bits, the model will suffer a significant accuracy drop. Following this motivation, we first analyze the data distribution of elements in the KV cache in Section~\ref{distribution} and find that these elements follow a normal distribution in token dimensions and even within each block. Based on this observation, we propose in Section~\ref{NQKV} to use data types that conform to the normal distribution to quantize at the block granularity, thereby minimizing quantization errors as much as possible. To reduce the additional overhead caused by quantization, Section~\ref{padding} proposes a strategy to pad in token dimensions, thereby improving the efficiency of quantization and dequantization operations.

\subsection{Data Distribution in KV cache}
\label{distribution}

Nowadays, the weights of LLMs can be quantized to 4 bits or even lower with minimal impact on model performance \cite{frantar2022gptq}. However, quantizing activations remains a challenging task due to the presence of outliers \cite{xiao2023smoothquant} \cite{dettmers2024qlora}. Since the KV cache essentially stores activations generated during the model inference process, quantizing the KV cache is also affected by outliers. Therefore, observing the data distribution in the KV cache is necessary, as it can help us understand the difficulties in quantizing the KV cache.

We collected the KV cache generated by each layer of the OPT-6.7B model during the inference process. Random samples of tokens were selected from the KV cache of each layer to observe their data distribution. As shown in Fig.~\ref{fig.distribution}, the data within each token mostly conforms to a normal distribution, and the standardized data closely matches the probability density function curve of the standard normal distribution. Therefore, we can conclude that, for the KV cache, the activation values within each token follow a normal distribution.

\begin{table}
\centering
\caption{
Results of the D’Agostino-Pearson (DAP) test for the data within each block. When the p-value is greater than the significance level $\alpha=0.05$, we fail to reject the null hypothesis, indicating that the data follows a normal distribution. The DAP test results for most blocks showed p-values much greater than the significance level $\alpha=0.05$, hence indicating that the data within each block follows a normal distribution.}
\label{table:test}
        \begin{tabular}{llllll}
        
        \toprule
        
        \textbf{block}  & pvalue & $>\alpha$? & \textbf{block}  & pvalue & $>\alpha$? \\   
        
        \midrule

        \textbf{0} & 0.61048 & \textcolor{green}{\ding{51}} & \textbf{8}  & 0.79392 & \textcolor{green}{\ding{51}} \\ 
        \textbf{1} & 0.19510 & \textcolor{green}{\ding{51}} & \textbf{9}  & 0.89790 & \textcolor{green}{\ding{51}} \\ 
        \textbf{2} & 0.26376 & \textcolor{green}{\ding{51}} & \textbf{10} & 0.74527 & \textcolor{green}{\ding{51}} \\ 
        \textbf{3} & 0.57718 & \textcolor{green}{\ding{51}} & \textbf{11} & 0.08653 & \textcolor{green}{\ding{51}} \\ 
        \textbf{4} & 0.32071 & \textcolor{green}{\ding{51}} & \textbf{12} & 0.71710 & \textcolor{green}{\ding{51}} \\ 
        \textbf{5} & 0.97007 & \textcolor{green}{\ding{51}} & \textbf{13} & 0.16879 & \textcolor{green}{\ding{51}} \\ 
        \textbf{6} & 0.51170 & \textcolor{green}{\ding{51}} & \textbf{14} & 0.59332 & \textcolor{green}{\ding{51}} \\ 
        \textbf{7} & 0.10981 & \textcolor{green}{\ding{51}} & \textbf{15} & 0.14138 & \textcolor{green}{\ding{51}} \\ 
        \bottomrule
\end{tabular}
\end{table}

In addition, we divided each token into several blocks using a fixed block size and explored the data distribution within each block. Q-Q (Quantile-Quantile) plots were created separately for the data distribution within tokens and within blocks. In statistics, a Quantile-Quantile plot is a probability plot, a graphical method for comparing two probability distributions by plotting their quantiles against each other \cite{gnanadesikan1968probability}. If the data points are as close as possible to the identity line $y = x$, it indicates that the data conforms to a standard normal distribution. As shown in Fig.~\ref{fig.qq.token}, the data within tokens follow a normal distribution. When the block size is set to 256, each token in the OPT-6.7B model is divided into 16 blocks. As shown in Fig.~\ref{fig.qq.block1}, \ref{fig.qq.block2}, and \ref{fig.qq.block3}, the data points within each block in the Q-Q plot approximately lie on the identity line $y = x$. Therefore, the data within blocks also conforms to a normal distribution. In Table~\ref{table:test}, we also conducted the D'Agostino-Pearson(DAP) test~\cite{d1986tests} to further test the normality of the data within each block. D'Agostino-Pearson (DAP) test is a statistical test used to determine whether a given sample of data comes from a normally distributed population. The null hypothesis for the D'Agostino-Pearson test is that the data follows a normal distribution. By performing the test and comparing the p-value to a significance level $\alpha=0.05$, we can determine whether to reject the null hypothesis or not. For blocks within each token, the p-value is much greater than the significance level $\alpha$, so we fail to reject the null hypothesis, \emph{suggesting that the data within each block follows a normal distribution.}

\begin{figure}[t]
    \centering
    \includegraphics[width=1.05\linewidth]{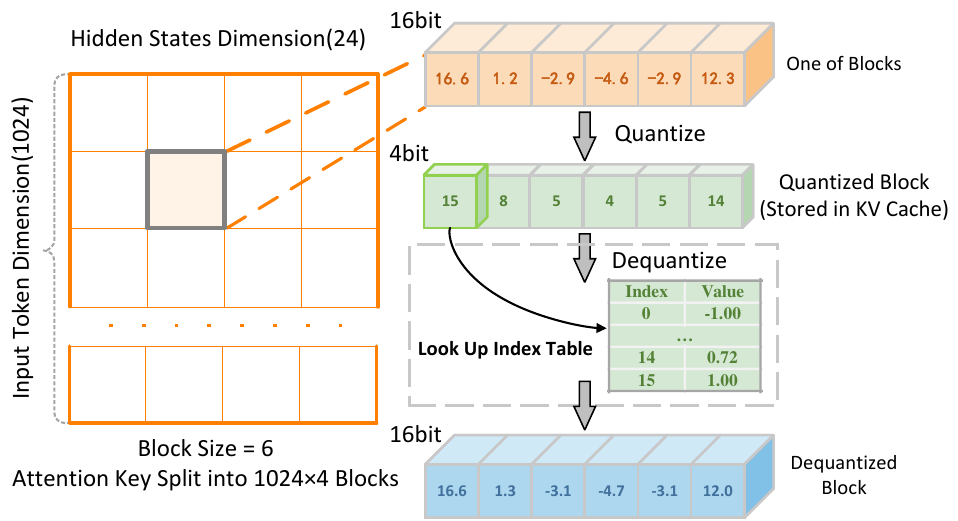}
    \caption{Block-wise quantile quantization. 
For demonstration purposes, let's assume the hidden states size is 24, input token dimension size is 1024, the block size is 6, and the dimensions of the keys matrix are 1024×24 (ignoring batch size). Therefore, each token of the keys can be divided into 4 blocks, and a keys matrix has 1024×4 blocks. We quantize each block separately, obtaining NF4 indices after quantization, which are stored in the KV cache. During dequantization processs, the NF4 indices stored in the KV cache can be used to look up the index table and get corresponding values, which are then restored to FP16 data type for computation.}
    \label{fig.block}
\end{figure}

\subsection{NQKV Algorithm}
\label{NQKV}

As we previously analyzed, the data within each block of the KV cache follows a normal distribution. Based on this observation, we propose a novel KV cache quantization approach called NQKV. The main idea of this approach is to partition the KV cache into blocks and use data types that conform to a normal distribution, such as Normal Float \cite{dettmers2024qlora}, for quantization of each block. In NQKV, we employ 4-bit Normal Float (NF4) \cite{dettmers2024qlora} data type for block-wise quantization, as shown in Fig.\ref{fig.block}.

The LLM attention inference process can be divided into two phases: the prefill phase and the decoding phase. In the prefill phase, the input prompt is used to generate keys and values for each transformer layer within LLMs. NQKV divides the generated keys and values into blocks along the token dimension, and applies NF4 quantization to each block, storing the resulting indices in the KV cache. The NQKV algorithm stores indices in the KV cache rather than directly storing floating point numbers. Although indices cannot be directly used for computation, we can retrieve corresponding floating point values based on the indices through table lookup. Since indices are stored using 4 bits, this effectively reduces the number of bits required to store the KV cache, saving approximately four times more memory compared to directly storing 16-bit floating-point numbers. Subsequently, in the decoding phase, newly generated keys and values are first quantized using per-block NF4 quantization and directly appended to the end of the KV cache, aligning with the streaming nature of the KV cache. Then, the KV cache is dequantized, and the resulting tensors are directly used in the subsequent computation of the attention mechanism. More specifically, we formalize the NQKV algorithm as the following process, which is also illustrated in Fig.\ref{fig.NQKV}:

\begin{figure*}
    \centering
    \includegraphics[width=0.85\textwidth]{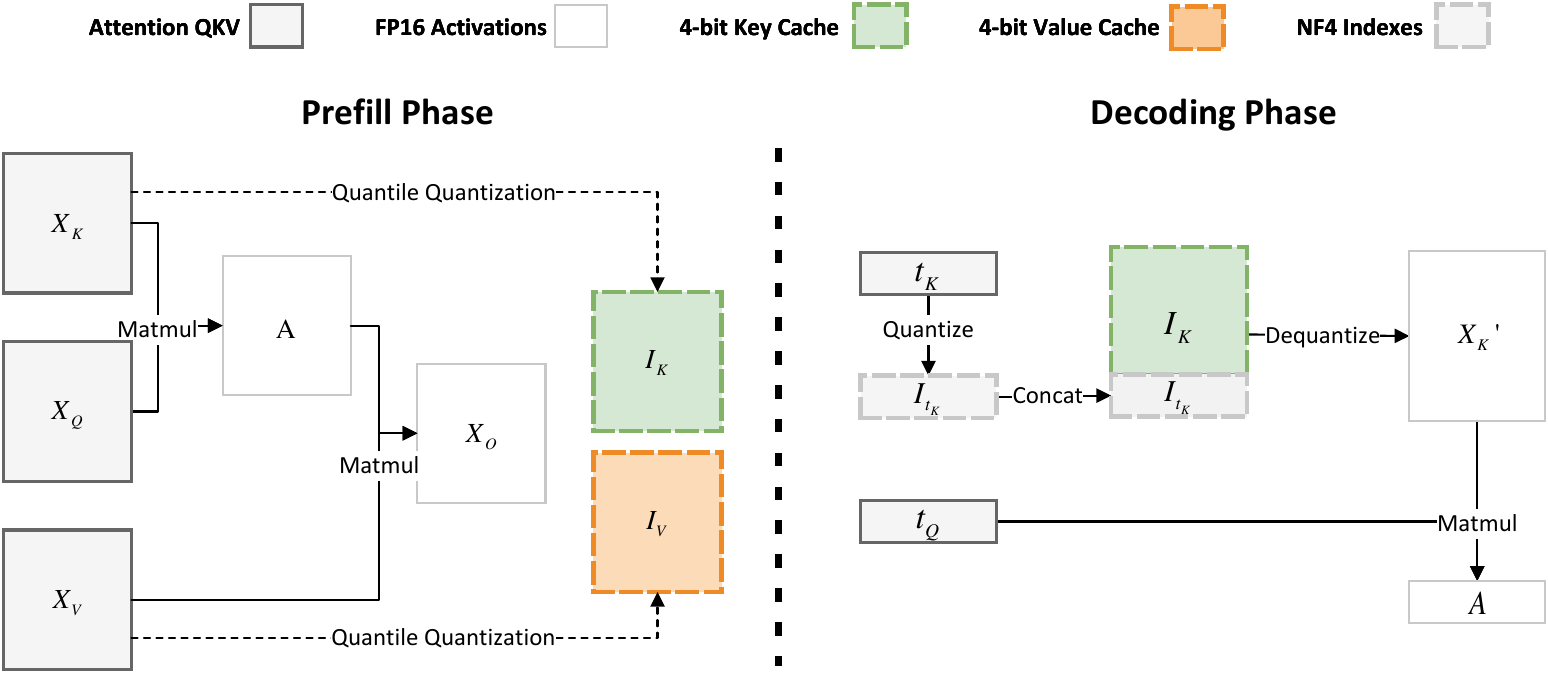}
    \caption{Execution flow of the NQKV algorithm. For ease of description, only the scenario of Key cache is described in the decoding phase, with the situation for Value cache being identical.}
    \label{fig.NQKV}
\end{figure*}
\textbf{Prefill Phase.} Let $X \in R^{b\times l_{prompt} \times d}$, where $b$ is the batch size, $l_{prompt}$ is the length of the input prompt, and $d$ is the size of hidden states. 
$X_Q$, $X_K$, and $X_V$ are the query, key, and value in the attention mechanism, respectively, and they are calculated by the following formulas: 
$$
X_Q = XW_Q, \quad X_K = XW_K, \quad X_V = XW_V,
$$
$W_Q, W_K, W_V \in R^{d \times d}$are the query, key, and value layer weights in the attention mechanism, respectively. Let $I_K$, $I_V$ be the indices obtained after NF4 quantization, and they satisfy:
$$I_K = quantize_{NF4}(X_K)$$
$$I_V = quantize_{NF4}(X_V)$$
where $quantize_{NF4}$ represents the block-wise NF4 quantization operation, as shown in Fig.\ref{fig.block}. $I_K, I_V$ are stored in the KV cache to avoid redundant computation during the decoding phase.

\textbf{Decoding Phase.} Let $t \in R^{b \times 1 \times d}$ be the newly generated input token embeding, and $t_K = tW_K$ and $t_V = tW_V$ be the newly generated key and value, respectively. We first perform NF4 block-wise quantization on $t_K$ and $t_V$:
% \begin{equation*}
% I_{t_K} = quantize_{NF4}(t_K),
% \end{equation*}
% \begin{equation*}
% I_{t_V} = quantize_{NF4}(t_V),
% \end{equation*}
$$I_{t_K} = quantize_{NF4}(t_K),$$
$$I_{t_V} = quantize_{NF4}(t_V),$$
Where $I_{t_K}$ and $I_{t_V}$ are the 4-bit indices obtained after quantization. Then, we update the KV cache by directly appending $I_{t_K}$ and $I_{t_V}$ to the end of the KV cache:
$$I_K \leftarrow Concat(I_K, I_{t_K}),$$
$$I_V \leftarrow Concat(I_V, I_{t_V}),$$
Finally, since indices cannot be directly used for computation, we need to dequantize the KV cache to obtain floating point numbers for subsequent attention computation:
$$X_K' = dequantize_{NF4}(I_K)$$
$$X_V' = dequantize_{NF4}(I_V)$$
$$t_Q = tW_Q$$
$$A = Softmax(t_Q X_K'^T),$$
$$t_O = A X_V' $$
where $t_O$ is the output of the attention, 
$t$ is the new token generated from the previous inference, and $t_Q$ is the attention query of this token. For the ease of illustration, we ignore the other part of the decoder layer.

\begin{figure}[b]
    \centering
    \includegraphics[width=1.0\linewidth]{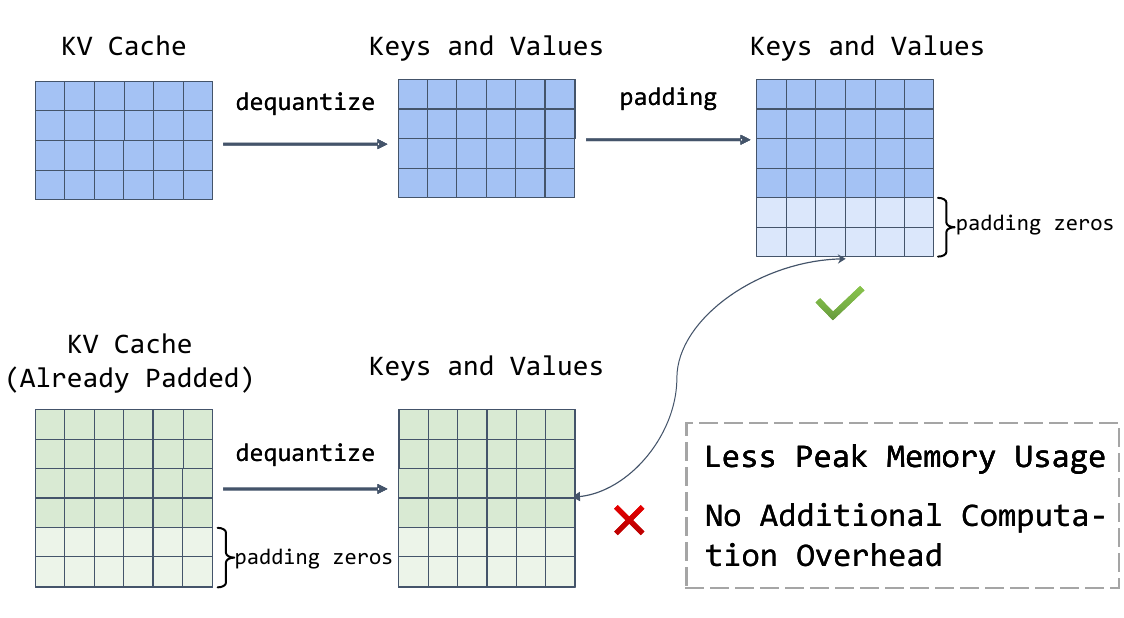}
    \caption{Padding the KV cache during computation would result in lower peak memory usage and has no additional computation overhead compared to directly padding the KV cache.}
    \label{fig4}
\end{figure}

\subsection{Padding}
\label{padding}

Our implementation of the NQKV algorithm is based on Nvidia's Cutlass template library. To leverage the GEMM (General Matrix Multiply) functionality provided by Cutlass and efficiently perform matrix multiplication operations, we employ padding techniques for the KV cache. Prior to computation, padding is applied to the KV cache along the token dimension to ensure that the token dimension size is a multiple of 16, meeting both GPU hardware requirements and optimization considerations.

As illustrated in Fig.~\ref{fig4}, we can apply padding directly to the KV cache, or we can perform padding after dequantizing the KV cache into computational values. The latter approach is indeed more efficient. Firstly, if padding is applied directly to the KV cache, the newly added elements will incur additional computational overhead during the dequantization process. Secondly, given the large number of layers in LLMs and the presence of KV caches in each layer, direct padding of the KV cache would result in additional storage overhead for each layer. However, during inference time, only one layer is active at any given time. Therefore, performing padding during the computation phase will only incur additional storage overhead for the KV cache of that single layer rather than each layer.

\begin{table*}[width=1.9\linewidth]
\centering
\caption{The impact of NQKV on the accuracy of the OPT models across different zero-shot tasks. NQKV has almost no impact on the accuracy of the OPT models, despite using a KV cache stored with only 4 bits.}

\label{table:base}
    \begin{tabular}{lcccccccc}
    % \multicolumn{9}{l}{\small{\textbf{Table 4}}}\\
    % \multicolumn{9}{l}{\small{Solution of our algorithm without trade credit}}\\
        \toprule
    
            Model & ~ & PIQA & WinoGrande & HellaSwag & ARC(Challenge) & RTE & boolq & \textbf{Average} \\
            
            \midrule
            
            \multirow{2}{*}{OPT-125M} & FP16 & 63.00\% & 50.36\% & 29.18\% & 19.11\% & 49.82\% & 55.47\% & 44.49\% \\ 
            ~ & NQKV & 62.68\% & 49.88\% & 28.94\% & 18.96\% & 51.26\% & 55.96\% & 44.61\% \\ 
    
            \hline
            
            \multirow{2}{*}{OPT-1.3B} & FP16 & 71.55\% & 59.51\% & 41.51\% & 23.38\% & 51.99\% & 57.77\% & 50.95\% \\ 
            ~ & NQKV & 71.07\% & 58.33\% & 40.48\% & 23.29\% & 51.62\% & 56.75\% & 50.26\% \\ 
    
            \hline
            
            \multirow{2}{*}{OPT-6.7B} & FP16 & 76.22\% & 65.35\% & 50.50\% & 30.63\% & 55.23\% & 66.06\% & 57.33\% \\ 
            ~ & NQKV & 76.17\% & 64.25\% & 50.16\% & 30.38\% & 55.60\% & 65.63\% & 57.03\% \\ 
    
            \hline
            
            \multirow{2}{*}{OPT-13B} & FP16 & 75.95\% & 65.04\% & 52.45\% & 32.94\% & 58.12\% & 65.93\% & 58.41\% \\ 
            ~ & NQKV & 75.84\% & 65.19\% & 52.14\% & 32.68\% & 59.21\% & 66.91\% & 58.66\% \\ 
            
    \bottomrule
    \end{tabular}
\end{table*}

\section{Experiment}
\subsection{Settings}

\textbf{Baselines.} 
To demonstrate the orthogonality of NQKV with other state-of-the-art quantization methods, we apply NQKV to SmoothQuant and test its impact on the accuracy of SmoothQuant. In our configuration for SmoothQuant, we quantize the weights at the per-tensor granularity and perform static per-tensor quantization for activations, i.e., scaling factors are computed and determined during the calibration phase and remain static during inference. We randomly select 512 sentences from the validation set of the Pile dataset \cite{gao2020pile} to generate scaling factors for activations, using a migration strength of $\alpha = 0.5$.

\textbf{Models and Datasets.} We evaluate NQKV using OPT \cite{zhang2022opt} model families. The OPT model is a decoder-only architecture based on the transformer's multi-head attention mechanism. We implemented the NQKV algorithm based on the Hugging Face\cite{jain2022hugging} transformers codebase. To achieve the best trade-off between accuracy and memory space occupation, we adopted 4-bit quantization with a block size of 256. We evaluated the model accuracy using seven zero-shot evaluation tasks, including PIQA \cite{bisk2020piqa}, WinoGrande \cite{sakaguchi2021winogrande}, HellaSwag \cite{zellers2019hellaswag}, ARC (Easy) \cite{clark2018arc}, ARC (Challenge) \cite{clark2018arc}, RTE \cite{wang2018glue}, and BoolQ \cite{devlin2018bert}. We utilized the lm-eval-harness\footnote{https://github.com/EleutherAI/lm-evaluation-harness} to evaluate OPT models ranging from 125M to 30B parameters. The experiments were conducted on a server equipped with 1 Nvidia A100 GPU (80GB).

\subsection{Accuracy Analysis}
\subsubsection{Accuracy on Zero-Shot Tasks}

To demonstrate that applying 4-bit Normal Float quantization to the KV cache only results in negligible accuracy degradation, we applied the NQKV method to the inference process of OPT models of various scales and evaluated their performance on various zero-shot tasks. We enabled the KV cache mechanism of the OPT model and applied 4-bit Normal Float quantization only to the keys and values within the multi-head attention mechanism of the OPT model during the inference process, while other activation values remained represented in the form of 16-bit floating point numbers. To demonstrate the impact of NQKV on model prediction performance, we did not apply any quantization strategy to the weights to avoid interference with the results.

Similar to SmoothQuant \cite{xiao2023smoothquant} and RPTQ \cite{yuan2023rptq}, we evaluated the accuracy on zero-shot tasks, and the results are shown in Table~\ref{table:base}. We observed that NQKV had almost no impact on the accuracy of the OPT model, despite the KV cache being stored with only 4 bits. Furthermore, as the model scale increased, the robustness of the LLM improved, and this impact became even smaller. Specifically, OPT-1.3B suffered an average accuracy loss of 0.7\%, but this loss was further reduced in the OPT-6.7B and OPT-13B models.

\subsubsection{Orthogonality to Other Methods}

To lower the barrier for deploying large models and accelerate inference, there are many advanced quantization methods available, such as SmoothQuant \cite{xiao2023smoothquant}, GPTQ \cite{frantar2022gptq}, and others. Since NQKV is specifically designed for quantizing KV cache, it does not conflict with existing advanced weight and activation quantization methods; they are orthogonal and can be used in combination. To demonstrate orthogonality, we applied the NQKV algorithm to SmoothQuant \cite{xiao2023smoothquant}, further quantizing KV cache to 4 bits based on SmoothQuant's W8A8 quantization. SmoothQuant offers various quantization granularities, and here we chose to use per-tensor quantization for both weights and activations. 
% SmoothQuant itself does not support inference using KV cache, so we made corresponding modifications to its source code to support inference acceleration using KV cache.

Table~\ref{table:orthogonality} indicates that the NQKV method has only a minor impact on the prediction accuracy of SmoothQuant, and in some cases, its prediction accuracy is even slightly higher than that of SmoothQuant. On the OPT-1.3B model, NQKV caused a relatively noticeable performance drop for SmoothQuant. However, as the model size increases, the robustness of large language models also improves. On the OPT-6.7B and OPT-30B models, NQKV actually brought an improvement in accuracy for SmoothQuant. Especially on the OPT-30B model, NQKV achieved higher accuracy than SmoothQuant on tasks such as WinoGrande, ARC(Easy), RTE, and BoolQ, with accuracy drop controlled within 0.1\% on other tasks. Our experiments show that, NQKV can work well in conjunction with SmoothQuant on large language models without causing catastrophic performance degradation.

\begin{table*}[width=2.0\linewidth]
\centering
\caption{Accuracy comparison on zero-shot tasks when combining NQKV with SmoothQuant. SQ represents the performance of original SmoothQuant algorithm, where weights, activations, and KV cache are all quantized to 8 bits(W8A8KV8). SQ-NQKV represents the performance of SmoothQuant combining with NQKV, where the KV cache is further quantized to 4 bits(W8A8KV4). NQKV demonstrates good orthogonality with other advanced quantization methods, as it does not cause catastrophic degradation in model performance. NQKV only incurs minimal accuracy drops, particularly on larger models. }
\label{table:orthogonality}
\scalebox{0.9}{
    \begin{tabular}{lccccccccc}
    % \multicolumn{10}{l}{\small{\textbf{Table 4}}}\\
    % \multicolumn{10}{l}{\small{Solution of our algorithm without trade credit}}\\
        \toprule
        
            Model & ~ & PIQA & WinoGrande & HellaSwag & ARC\newline(Easy) & ARC\newline(Challenge) & RTE & boolq & \textbf{Average} \\ 
    
        \midrule
                    
            ~ & FP16 & 63.00\% & 50.36\% & 29.18\% & 43.52\% & 19.11\% & 49.82\% & 55.47\% & 44.35\% \\ 
            OPT-125M & SQ & 62.46\% & 51.30\% & 28.85\% & 41.96\% & 19.28\% & 49.82\% & 56.21\% & 44.27\% \\ 
            ~ & SQ-NQKV & 62.24\% & 50.67\% & 28.63\% & 42.59\% & 18.94\% & 50.18\% & 56.36\% & 44.23\% \\ 
    
         \hline
            
            ~ & FP16 & 71.55\% & 59.51\% & 41.51\% & 57.11\% & 23.38\% & 51.99\% & 57.77\% & 51.83\% \\ 
            OPT-1.3B & SQ & 70.40\% & 58.72\% & 41.26\% & 56.65\% & 24.40\% & 51.26\% & 56.54\% & 51.32\% \\ 
            ~ & SQ-NQKV & 70.24\% & 59.27\% & 40.16\% & 53.62\% & 23.29\% & 50.18\% & 55.35\% & 50.30\% \\ 
    
         \hline
            
            ~ & FP16 & 76.22\% & 65.35\% & 50.50\% & 65.66\% & 30.63\% & 55.23\% & 66.06\% & 58.52\% \\ 
            OPT-6.7B & SQ & 76.50\% & 65.82\% & 50.42\% & 65.49\% & 30.08\% & 55.60\% & 66.33\% & 58.61\% \\ 
            ~ & SQ-NQKV & 76.44\% & 66.61\% & 50.08\% & 65.28\% & 29.69\% & 56.68\% & 66.06\% & 58.69\% \\ 
    
        \hline
            
            ~ & FP16 & 75.95\% & 65.04\% & 52.45\% & 67.13\% & 32.94\% & 58.12\% & 65.93\% & 59.65\% \\ 
            OPT-13B & SQ & 75.68\% & 64.56\% & 52.15\% & 66.75\% & 33.11\% & 57.40\% & 64.65\% & 59.19\% \\ 
            ~ & SQ-NQKV & 75.73\% & 65.11\% & 51.70\% & 65.70\% & 32.25\% & 55.60\% & 64.56\% & 58.66\% \\ 
    
        \hline
            
            ~ & FP16 & 77.64\% & 68.35\% & 54.30\% & 70.12\% & 34.56\% & 57.76\% & 70.49\% & 61.89\% \\ 
            OPT-30B & SQ & 77.53\% & 67.64\% & 54.04\% & 69.99\% & 34.39\% & 56.68\% & 69.94\% & 61.46\% \\ 
            ~ & SQ-NQKV & 77.48\% & 67.88\% & 53.94\% & 70.16\% & 34.13\% & 58.84\% & 70.58\% & 61.86\% \\ 
        \bottomrule
    \end{tabular}
}
\end{table*}

\begin{figure}
    \centering
    \includegraphics[width=0.9\linewidth]{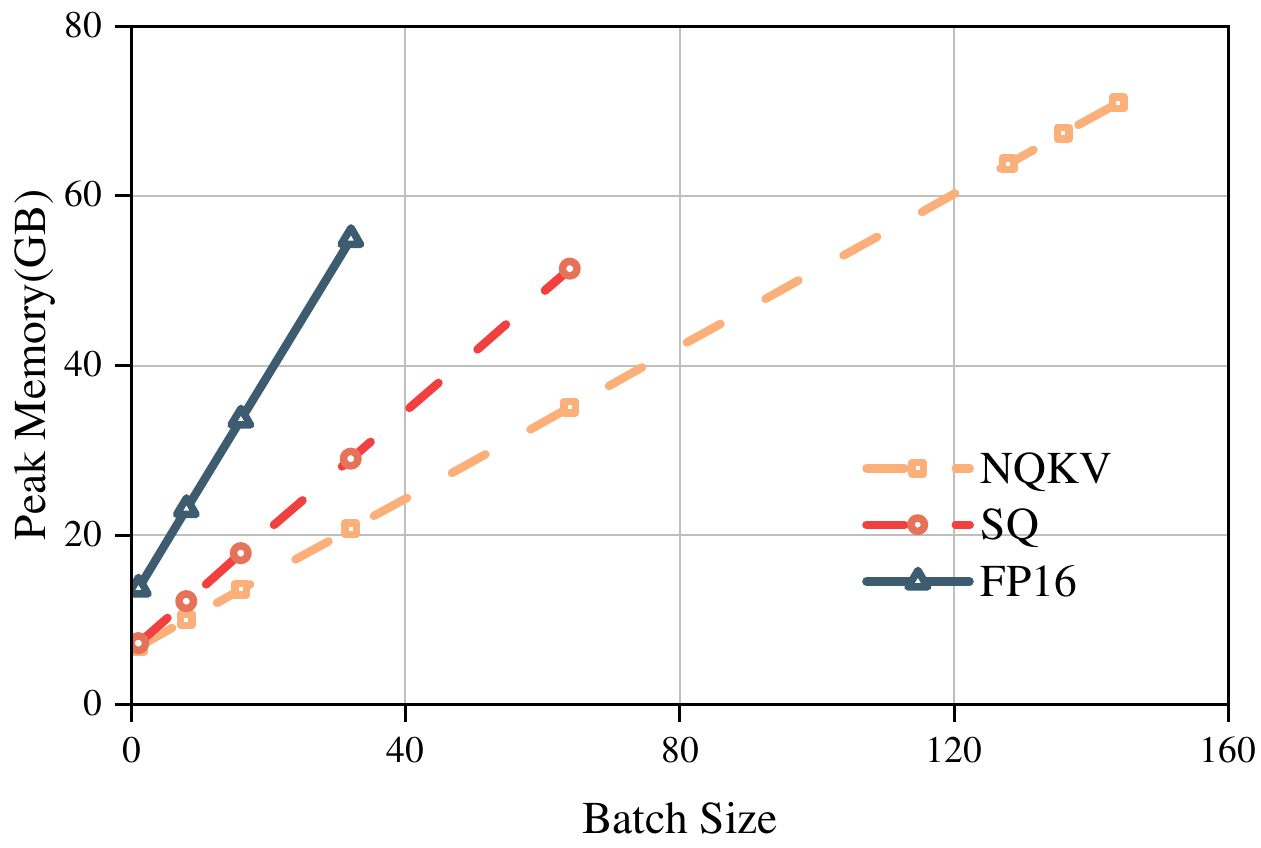}
    \caption{For the OPT-6.7B model, NQKV can perform inference with 4 $\times$ batch size compared to a standard FP16 model, and with 2 $\times$ batch size compared to SmoothQuant.}
    \label{fig.mem1}
\end{figure}

\begin{figure}
    \centering
    \includegraphics[width=0.91\linewidth]{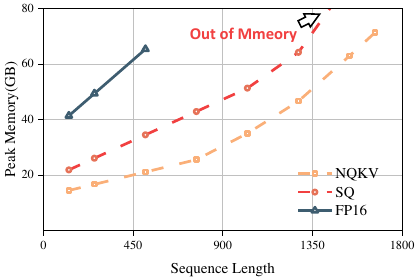}
    \caption{For the OPT-6.7B model, NQKV can perform inference with 2.5 $\times$ sequence length compared to a standard FP16 model, and with 1.5 $\times$ sequence length compared to SmoothQuant.}
    \label{fig.mem2}
\end{figure}

\subsection{Speedup and Memory Saving}

To measure the impact of NQKV on the throughput and memory usage of LLMs with enabled KV cache, we use the wikitext-2 dataset as the workload for text generation tasks. The number of the input tokens of the model is determined by the sequence length, and the output length \( l_{gen} \) is 338. By varying the batch size and sequence length, we observe the performance of the OPT-6.7B and OPT-30B models under this workload. Here, our GPU is the Nvidia A100 GPU (80GB). We measure the throughput of OPT models and measure the peak memory usage during inference time as a metric for memory efficiency. 

In Fig.\ref{fig.mem1} and Fig.\ref{fig.mem2}, we show 
 that NQKV can save a significant amount of memory space, allowing for larger batch sizes or longer contexts for inference. For OPT-6.7B model, when the FP16 model cannot continue inference due to insufficient memory, NQKV allows the model to still perform inference with 4$\times$  batch size or 2.5$\times$ sequence length. For larger models, the memory saving effect of NQKV will be even more significant.

As shown in Table~\ref{table:throughput}, with the KV cache enabled, NQKV allows SmoothQuant to perform inference with an 2$\times$ larger batch size or a 4$\times$ longer sequence length, with a throughput loss of less than 20\%. For OPT-30B model, when the batch size is 64 and the sequence length is 512, SmoothQuant cannot enable the KV cache normally because it would result in an out of memory error. However, with NQKV, SmoothQuant can enable KV cache and perform inference at a speed 9.3 $\times$ faster, with only a 5\% increase in memory usage. Overall, NQKV can save an additional 60\%-80\% of memory compared to SmoothQuant when the batch size and sequence length are very large.

We observed that the throughput of NQKV is slightly lower than that of SmoothQuant. This is because NQKV uses storage types instead of computation types to store the quantized KV cache, resulting in additional overhead to dequantize the KV cache into computation values during calculations. Nonetheless, compared to scenarios without using KV cache, NQKV still provides a significant inference acceleration. For smaller models (such as OPT-6.7B), NQKV enables the use of KV cache almost without additional memory overhead. This means we can accelerate the inference of these models with nearly no extra memory cost incurred. For larger models, when other methods are unable to enable KV cache due to memory limitations, NQKV can still enable KV cache and achieve accelerated inference.

It's worth noting that in some cases (such as OPT-6.7B, with a batch size of 8 and sequence length of 512), the peak memory usage of NQKV may even be lower than that of SmoothQuant without enabling KV cache. This seems counterintuitive, as enabling KV cache would inevitably incur additional memory overhead, making it impossible to achieve a smaller peak memory footprint. In fact, this is because our implementation is based on PyTorch, and PyTorch's memory allocation strategy may allocate more memory than necessary for the model, leading to such results. The peak memory usage determines whether the model can perform inference on the GPU, so we use this metric instead of the average memory usage during inference.

\begin{table*}[width=1.9\linewidth]
\centering\small
\caption{Comparison of throughput and memory usage of OPT models under different configurations. BS represents Batch Size, SeqLen represents Sequence Length. SQ is SmoothQuant without using KV cache, SQKV is SmoothQuant with KV cache, and NQKV represents using NQKV algorithm to further quantizes the KV cache based on SQKV. OOM indicates out of memory errors.}
\label{table:throughput}
% \begin{tabular}{p{1.5cm} p{1cm} p{1.1cm} *{3}{p{1.4cm}<{\raggedright}} p{0.2cm} *{3}{p{1.4cm}<{\raggedright}}}
\begin{tabular}{lcccccccccc}
% \multicolumn{10}{l}{\small{\textbf{Table 4}}}\\
% \multicolumn{10}{l}{\small{Solution of our algorithm without trade credit}}\\
    \toprule

        % \multirow{2}{*}{Model} & \multirow{2}{*}{BS} & \multirow{2}{*}{SeqLen} & Throughput(token/s) & ~ & ~ & ~ & \mbox{Peak Mem(MB)} & ~ & ~ \\ 
        \multirow{2}{*}{Model} & \multirow{2}{*}{BS} & \multirow{2}{*}{SeqLen} & \multicolumn{4}{l}{Throughput(token/s)} & \multicolumn{4}{l}{\mbox{Peak Mem(GB)}} \\ 

        \cmidrule(lr){4-7}
        \cmidrule(lr){8-11}
                & & &  SQ & SQKV & NQKV & Speedup(↑) & SQ & SQKV & NQKV & Saving (↑) \\ 
        \midrule

        \multirow{6}{*}{OPT-6.7B} & \multirow{2}{*}{8} & 128 & 45.55 & 139.78 & 118.69 & 2.61 & 7.42 & 8.28 & 6.92 & 1.20 \\ 
        ~ & ~ & 512 & 18.17 & 103.94 & 91.38 & 5.03 & 8.95 & 9.84 & 7.91 & 1.24 \\ 
        \cmidrule(lr){2-11}
        ~ & \multirow{2}{*}{32} & 128 & 57.65 & 275.48 & 258.13 & 4.48 & 9.07 & 11.96 & 8.39 & 1.43 \\ 
        ~ & ~ & 512 & 19.63 & 151.97 & 131.86 & 6.72 & 14.41 & 18.28 & 12.41 & 1.47 \\ 
        \cmidrule(lr){2-11}
        ~ & \multirow{2}{*}{64} & 128 & 58.53 & 311.23 & 289.31 & 4.94 & 14.46 & 21.39 & 15.39 & 1.39 \\ 
        ~ & ~ & 512 & 20.61 & 172.28 & 153.64 & 7.45 & 26.72 & 33.73 & 18.40 & 1.83 \\ 

        \midrule
        
        \multirow{6}{*}{OPT-30B} & \multirow{2}{*}{8} & 128 & 14.07 & 85.02 & 70.84 & 5.03 & 29.97 & 33.43 & 29.76 & 1.12 \\ 
        ~ & ~ & 512 & 5.55 & 59.88 & 51.68 & 9.31 & 32.43 & 37.36 & 31.87 & 1.17 \\ 
        \cmidrule(lr){2-11}
        ~ & \multirow{2}{*}{32} & 128 & 19.66 & 97.14 & 93.26 & 4.74 & 32.00 & 42.19 & 33.50 & 1.26 \\ 
        ~ & ~ & 512 & 6.21 & 60.32 & 54.27 & 8.74 & 40.26 & 58.23 & 41.97 & 1.39 \\ 
        \cmidrule(lr){2-11}
        ~ & \multirow{2}{*}{64} & 128 & 18.36 & 112.79 & 98.91 & 5.39 & 40.45 & 67.50 & 41.53 & \textbf{1.63} \\ 
        ~ & ~ & 512 & 5.71 & -- & 53.10 & \textbf{9.30} & 60.12 & \textbf{OOM} & 62.49 & \textbf{--} \\ 
    \bottomrule
\end{tabular}
\end{table*}

\section{Conclusion and Future Work}

In this paper, we conducted an extensive analysis of the element distribution within the KV cache and found that both within individual tokens and within individual blocks, the elements follow a normal distribution. Based on this observation, we conclude that using data types whose data points follow a normal distribution for per-block quantization of KV cache can further reduce quantization errors. Furthermore, we propose the NQKV algorithm, an effective quantization method that specifically designed for KV cache and does not need any retraining or finetuning. Our experiments demonstrate that our method allows for an 2$\times$ larger batch size or 4$\times$ larger sequence length for inference when KV cache is enabled, and it improves throughput by 9.3$\times$ compared to the scenario without using KV cache. In the future, we will further optimize the implementation of NQKV to reduce the overhead of quantization on LLM inference. Additionally, we will explore the design of new data types in hopes of further reducing quantization errors.

\printcredits

\section*{Acknowledgements}
This research is supported by the National Natural Science Foundation of China (62372366).

\bibliographystyle{cas-model2-names}
%\bibliographystyle{unsrt}

\bibliography{cas-refs}

\vskip5pt

% \vskip55pt

\bio{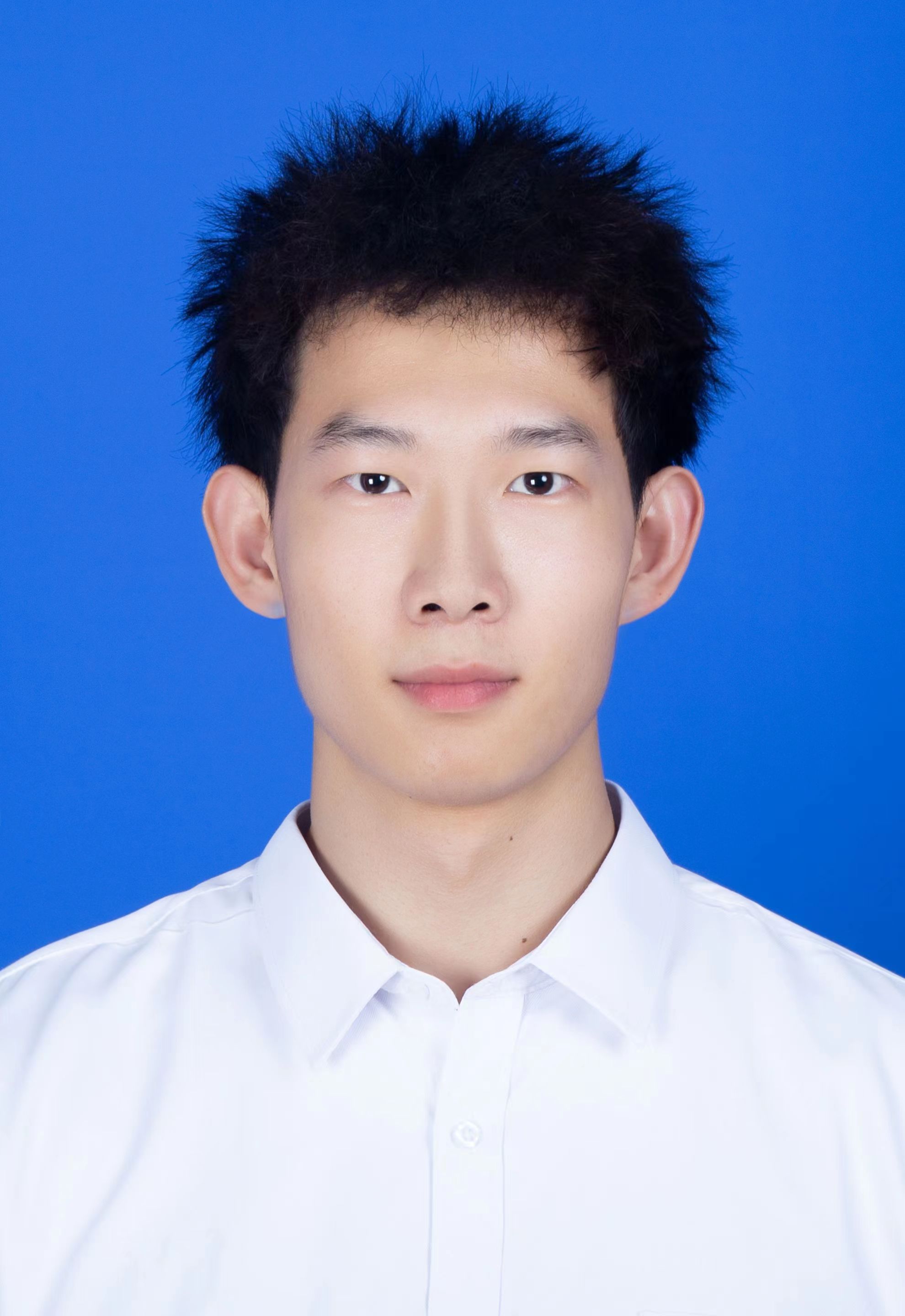}
Zhihang Cai received the B.S. degrees from the school of Xi'an Jiaotong University, Xi’an, China, in 2019 and 2023, respectively. He is currently pursuing a M.S. degree with Xi’an Jiaotong University, Xi’an, China. His research interests include machine learning and computer architecture.
\endbio

\vskip55pt

\bio{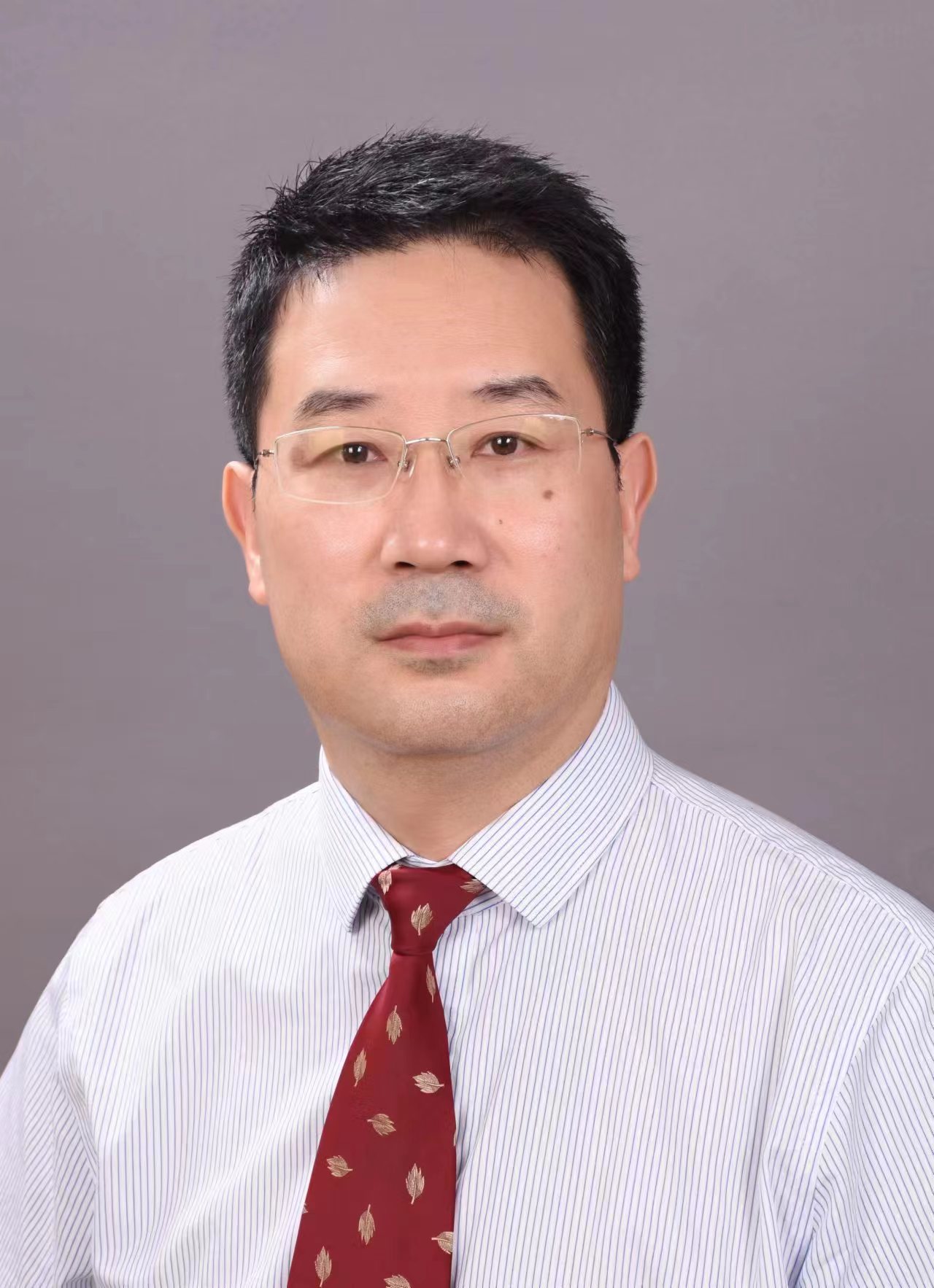}
Xingjun Zhang (Member, IEEE) received his Ph.D. degree in Computer Architecture from Xi’an Jiaotong University, China, in 2003. From Jan. 2004 to Dec. 2005, he was Postdoctoral Fellow at the Computer School of Beihang University, China.From Feb. 2006 to Jan. 2009, he was Research Fellow in the Department of Electronic Engineering of Aston University, United Kingdom. He is now a Full Professor and the Dean of the School of Computer Science \& Technology, Xi’an Jiaotong University. His research interests include high-performance computing, big data storage system, and distributed machine learning.
\endbio

\vskip5pt

\bio{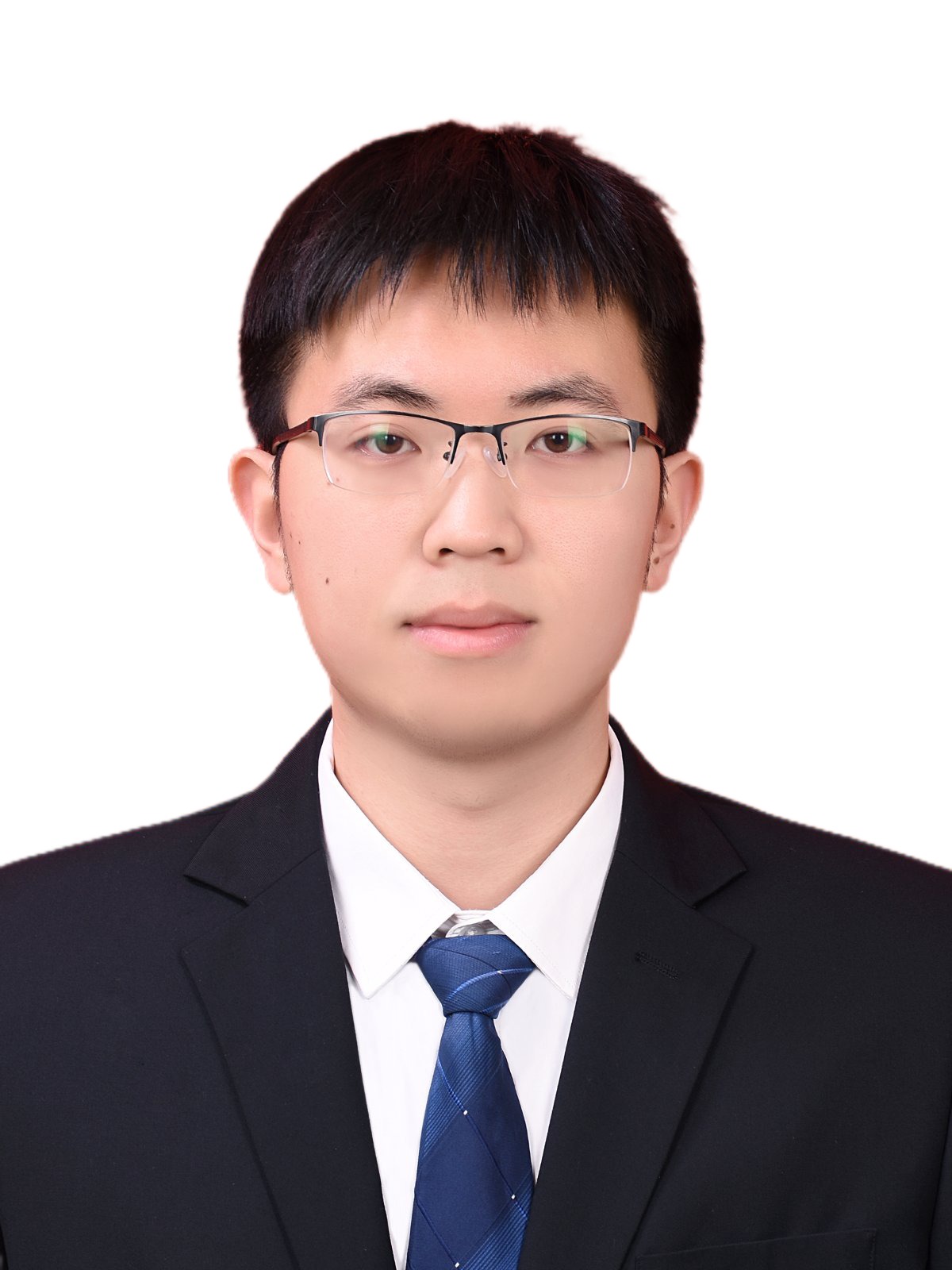}
Zhendong Tan received the B.S.  degrees from the school of Xi'an Jiaotong University, Xi’an, China, in 2019 and 2023, respectively. He is currently pursuing a Ph.D. degree with Xi’an Jiaotong University, Xi’an, China. His research interests include efficient machine learning and computer architecture.
\endbio

\vskip45pt

\bio{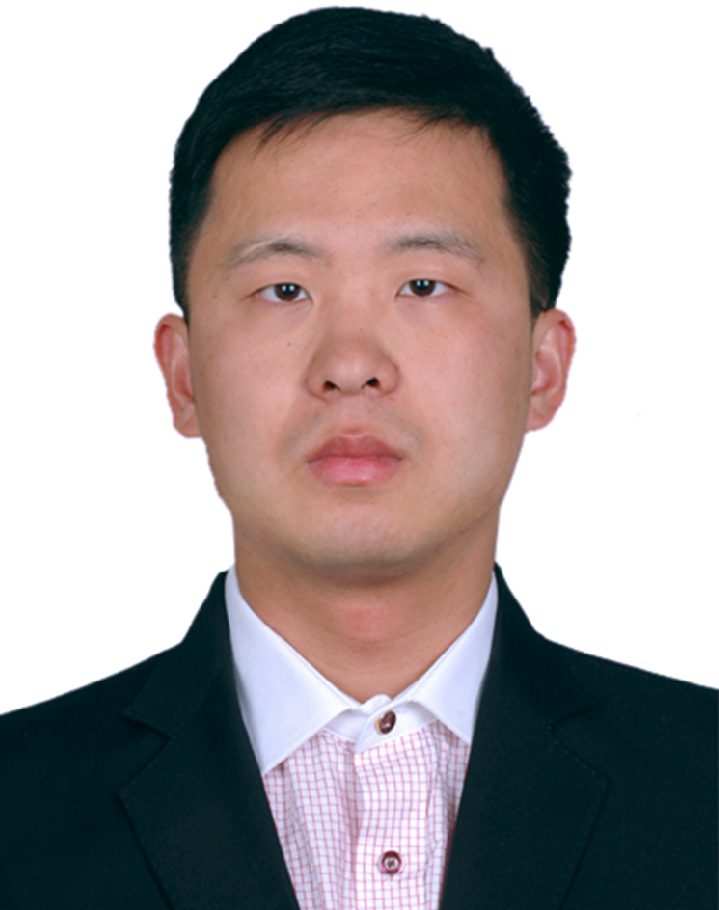}
Zheng Wei received the B.S. and M.S. degrees from the school of Communication Engineering from Xidian University, Xi’an, China, in 2013 and 2016, respectively. He is currently pursuing a Ph.D. degree with Xi’an Jiaotong University, Xi’an, China. His research interests include machine learning, computer architecture, and hardware accelerators for deep learning.
\endbio

\end{document}